\pdfoutput=1

\documentclass[11pt]{article}

\usepackage[]{acl}

\usepackage{times}
\usepackage{latexsym}
\usepackage{enumitem}
\usepackage{multirow}
\usepackage{amsmath}
\usepackage{booktabs}
\usepackage{subfigure}
\usepackage{diagbox}
\usepackage{amssymb}

\usepackage[T1]{fontenc}

\usepackage[utf8]{inputenc}

\usepackage{microtype}

\usepackage{graphicx}

\usepackage[most]{tcolorbox}
\usepackage{listings}
\usepackage{xcolor}
\newtcolorbox{promptbox}[1][]{
    colback=gray!5,
    colframe=black!70,
    fonttitle=\bfseries,
    coltitle=white,
    colbacktitle=black!70,
    boxrule=0.6pt,
    arc=2pt,
    left=6pt, right=6pt, top=4pt, bottom=4pt,
    breakable,
    enhanced,
    #1
}

\title{EvoBrowseComp: Benchmarking Search Agents on Evolving Knowledge}

\author{Yunhan Wang\textsuperscript{$\spadesuit$$\clubsuit$}\thanks{ \ \ Work was done when Yunhan Wang was interning at Weixin AI, Tencent Inc, China.}, Jiaan Wang\textsuperscript{$\spadesuit$}\footnotemark[2] \\ \textbf{Lianzhe Huang\textsuperscript{$\spadesuit$}, Xianfeng Zeng\textsuperscript{$\spadesuit$} and \ Fandong Meng\textsuperscript{$\spadesuit$}\thanks{ \ \ Corresponding authors.}} \\
\textsuperscript{$\spadesuit$}Weixin AI, Tencent Inc, China \ \ \textsuperscript{$\clubsuit$}Northeastern University, China \\ 
\texttt{yunhannnan@gmail.com}\ \   \texttt{\{torchwang,fandongmeng\}@tencent.com}
}

\begin{document}
\maketitle
\begin{abstract}
Search Agents---large language models augmented with search tools---have intensified the need for future-proof evaluation benchmarks. Existing benchmarks such as BrowseComp rely on static knowledge, making them vulnerable to test-set contamination and parametric memorization. Consequently, models can achieve high scores through fact recall rather than genuine retrieval, obscuring true browsing competence via reasoning shortcuts. 

In this paper, we introduce EvoBrowseComp, an \textit{evolving} benchmark of 400 English and 400 Chinese contamination-free complex questions synthesized via live-web traversal. To collect these questions, we design a three-agent collaborative framework: (1) a QA synthesis agent that retrieves \emph{fresh} knowledge from the live web to synthesize QA pairs; (2) an information filtering agent that filters retrieved knowledge in terms of credibility and popularity to block parametric shortcuts; and (3) a high-level guidance agent that formalizes questions into reasoning graphs to reduce logical redundancy and shortcuts in synthesized QA pairs.
Because the framework supports fully automated synthesis, EvoBrowseComp can be regularly updated to prevent data contamination and maintain temporal freshness.
Extensive experiments confirm its great difficulty, requiring broad horizontal search. It establishes a scalable paradigm for auto-updatable, high-difficulty benchmarking that keeps pace with both evolving world knowledge and advancing agent capabilities.\footnote{We have released the data at \url{https://hf.co/datasets/Krystalan/EvoBrowseComp}}
\end{abstract}

\section{Introduction}

\begin{figure}[!t]
\centering
\includegraphics[width=1.0\linewidth]{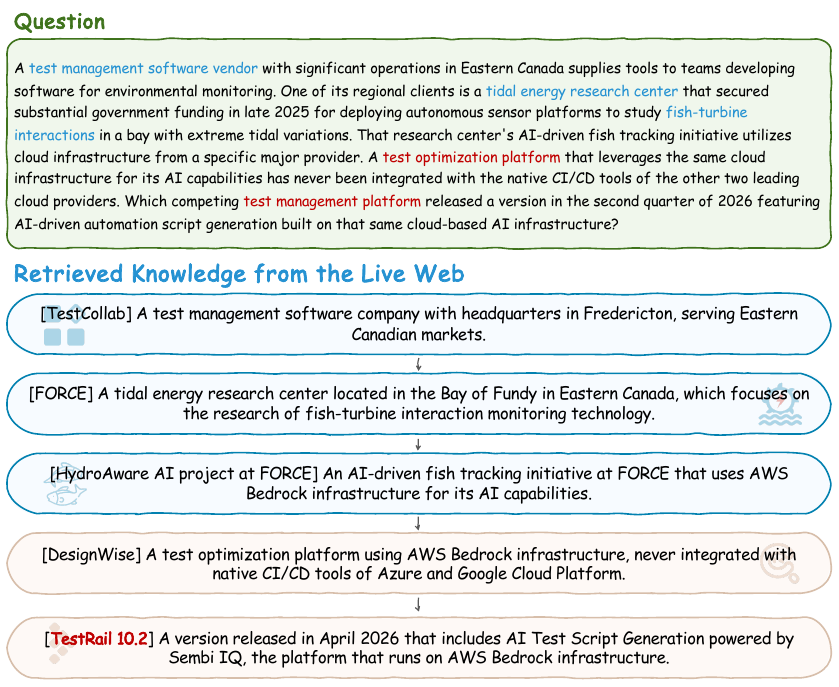}
\caption{An illustrative example question from EvoBrowseComp. \textcolor[RGB]{215,190,170}{Orange} highlights fresh knowledge (post-2026) lies in its reasoning paths, while \textcolor[RGB]{192,0,0}{red} denotes its final answer.}
\label{fig:example_question}
\end{figure}

Large Language Models (LLMs) augmented with web search tools, known as search agents~\cite{wei2025browsecomp,chen2025browsecomp,zhou2025browsecomp}, have demonstrated remarkable performance on information-seeking tasks. These agents exercise \textit{web browsing ability}---persistently navigating the open web, executing multi-hop questions, and gathering fragmented evidence across different sources~\cite{wu-etal-2025-webwalker, gupta2026deepsearchqa}.
To measure this ability, many benchmark datasets are proposed sequentially. BrowseComp~\cite{wei2025browsecomp} and BrowseComp-ZH~\cite{zhou2025browsecomp} focus on horizontal search, evaluating persistence and creativity in locating hard-to-find facts.
GAIA~\cite{mialon2024gaia} tests general assistant competence through real-world, multi-step tool use.
BFCL~\cite{patil2025the} assesses search orchestration via function calling, while WebWalker~\cite{wu-etal-2025-webwalker} isolates vertical traversal within structured websites.
More recently, specialized benchmarks have targeted higher-order retrieval competencies: SealQA~\cite{pham2025sealqa} probes robustness under noisy and conflicting retrieval conditions; while DeepSearchQA~\cite{gupta2026deepsearchqa} raises the bar by requiring exhaustive collation of answer sets across multiple sources.
These efforts have established a rich landscape for benchmarking LLMs' web browsing ability.

However, existing benchmarks are typically anchored to \textit{static} knowledge.
For example, BrowseComp~\cite{wei2025browsecomp} and BrowseComp-ZH~\cite{zhou2025browsecomp} relies on question--answer pairs manually curated at a fixed point in time; BrowseComp-Plus~\cite{chen2025browsecomp} freezes a curated document snapshot to ensure reproducibility; GAIA~\cite{mialon2024gaia} grounds its tasks in specific and immutable versions of web pages or attached files; and DeepSearchQA~\cite{gupta2026deepsearchqa}, though time-anchored, comprises a static prompt set evaluated against a fixed answer key. This static nature renders them acutely vulnerable to test-set contamination: as pre-training corpora expand, benchmark content inevitably leaks into model parameters, enabling models to solve questions via parametric memorization rather than genuine browsing and reasoning.
As pointed out by \citet{Anthropic2026EvalAwareness},  the explicit leakage of BrowseComp answers into public data confirms that this benchmark has been compromised by data contamination.

To address these limitations, we introduce \emph{EvoBrowseComp}, an evolving benchmark comprising 400 English and 400 Chinese complex questions automatically synthesized from live-web traversal. Our construction pipeline actively discovers and validates \textit{fresh} knowledge and automatically constructs QA pairs through a three-agent collaborative framework. First, a \textit{QA Synthesis Agent} retrieves fresh knowledge via web tools and provides (candidate) QA pairs based on the knowledge. Second, an \textit{Information Filtering Agent} filters out retrieved knowledge in terms of credibility (verifying source credibility and cross-source consistency) and popularity (blocking parametric shortcuts through over-exposed knowledge).
Third, a \textit{High-level Guidance Agent} structures each question as a reasoning graph using three basic operations: projection, intersection, and complement. It identifies both structural redundancies and shortcuts, and directs the QA synthesis agent toward specific synthesis directions.
Moreover, we adopt several strategies to ensure data quality, including the verification of textual quality, answer uniqueness, and question difficulty.
In this manner, high-quality challenging questions involving fresh knowledge can be automatically collected (c.f., Figure~\ref{fig:example_question}).
In contrast, prior benchmarks~\cite{mialon2024gaia,wei2025browsecomp,zhou2025browsecomp,chen2025browsecomp,pham2025sealqa,gupta2026deepsearchqa} typically rely on labor-intensive human curation that makes regular updates prohibitively expensive.
EvoBrowseComp removes this barrier: its synthesis pipeline is fully automated, requiring no costly manual annotation.
This enables the benchmark to be refreshed continuously at minimal cost, swapping in newly emerged facts while retiring over-exposed ones.

Based on EvoBrowseComp, we evaluate various LLMs under both tool-based and tool-free settings. The results reveal two critical phenomena. First, even Claude-Opus-4.8~\cite{Claude4.8}, a cutting-edge reasoning LLM, achieves only 46.2\% accuracy when equipped with tools, indicating that our temporally fresh, structurally complex questions are not easily retrieved.
Second, when tool access is removed, Claude-Opus-4.8's performance drops to 9.0\%, confirming that answering these questions demands genuine retrieval and multi-hop reasoning over fresh knowledge, rather than static recall.
We believe this establishes a sustainable, contamination-resistant paradigm for future-proof evaluation of search agents.

In summary, our contributions are as follows:
\begin{itemize}[leftmargin=*,topsep=0pt]
\setlength{\itemsep}{0pt}
\setlength{\parsep}{0pt}
\setlength{\parskip}{0pt}
    
\item We introduce EvoBrowseComp, a search agent benchmark comprising 400 English and 400 Chinese complex questions. Grounding questions in fresh knowledge, it prevents models from exploiting parametric memorization.
    
\item We propose a fully automated three-agent synthesis framework. It requires no costly human annotation, enabling continuous, low-cost regeneration that retires over-exposed questions and incorporates newly emerged facts and knowledge.
    
\item Extensive evaluations demonstrate that even frontier LLMs achieve only modest accuracy ($<$50\%) with web tools, and their performance collapses sharply when tool access is removed ($<$11\%). This confirms that EvoBrowseComp effectively isolates genuine web browsing and multi-hop reasoning from static parametric recall.
\end{itemize}

\section{EvoBrowseComp}

\begin{figure*}[t]
\centerline{\includegraphics[width=0.90\textwidth]{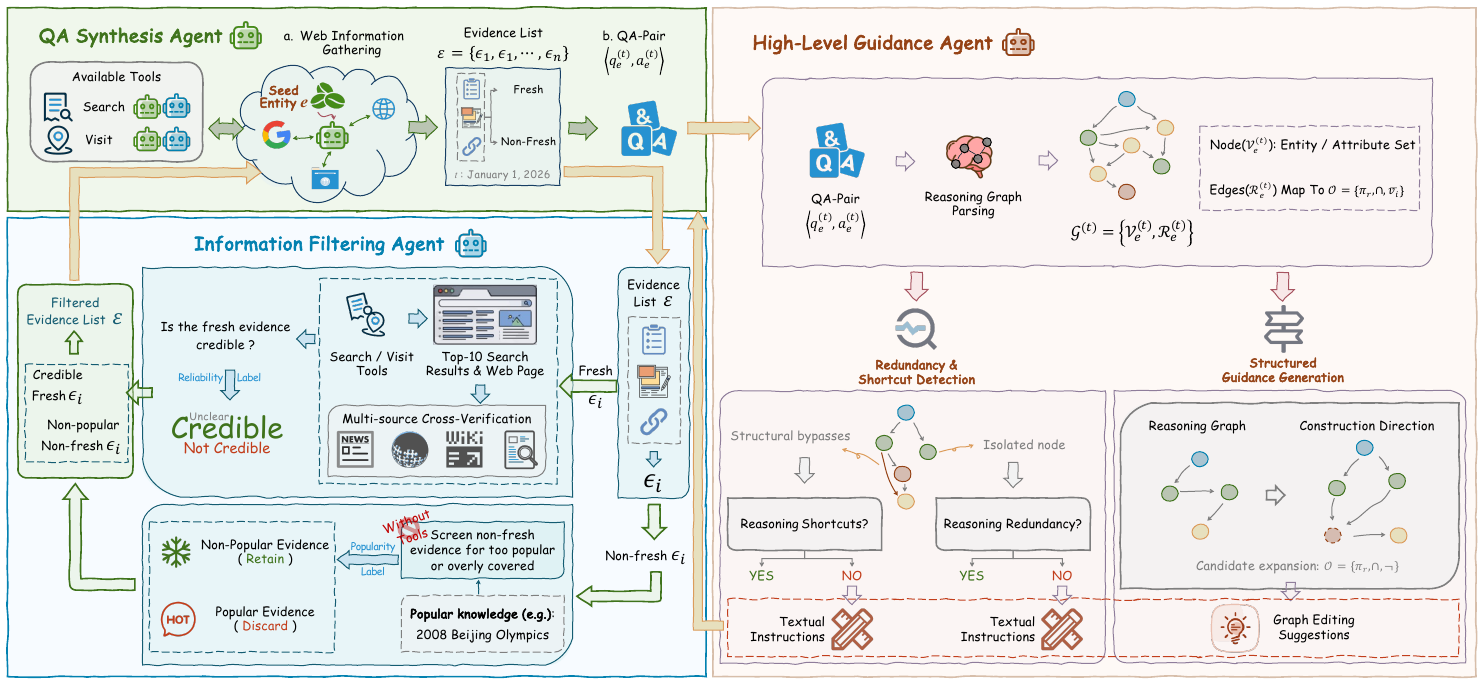}}
\caption{The illustration of the three-agent collaborative framework. (a) QA synthesis agent retrieves knowledge from the live web and generates (candidate) QA pairs; (b) Information filtering agent judges each retrieved knowledge in terms of credibility and popularity (popular/over-covered or not credible knowledge will be discarded); (c) High-level guidance agent detects the logical redundancy and shortcuts in the candidate QA pairs based on the constructed reasoning graphs, and gives suggestions to the QA synthesis agent in the next iteration.}
\label{fig:framework}
\end{figure*}

EvoBrowseComp is built on two foundational principles. First, \emph{questions should involve fresh knowledge}. By synthesizing questions from knowledge that emerges after training cutoffs, we prevent models from answering via parametric memorization. Second, \emph{the construction pipeline should be fully automated and continuously evolvable}. This enables periodic regeneration in which over-exposed questions are retired and replaced by newly surfaced knowledge, guaranteeing long-term benchmark validity without expensive human curation.

\subsection{Data Collection}
\label{subsec:2.1}

The data collection pipeline operates as an \textit{iterative feedback loop} among three specialized agents (c.f., Figure~\ref{fig:framework}).
Beginning with seed entities, a QA synthesis agent searches the live web to propose candidate QA pairs with retrieved knowledge.
Each retrieved knowledge is evaluated by an information filtering agent in terms of credibility and popularity.
A high-level guidance agent formalizes the underlying reasoning structure of a candidate question generated in the $i$-th iteration, detects its logical redundancy and shortcuts, and guides the QA synthesis agent in the next iteration.
In this way, the three agents collaborate automatically to synthesize highly complex, high-quality QA pairs.

\paragraph{Seed Entity.}
Synthesizing temporally fresh and logically complex QA pairs requires seed entities that tend to involve fresh knowledge. Rather than harvesting entities from a static knowledge graph---which risks stale facts---we collect seed entities through live-web retrieval.
Specifically, we pre-define 9 core domains (\emph{e.g.}, science, economy and geography) and 50 fine-grained sub-domains. For each sub-domain, we equip an advanced LLM, \emph{i.e.}, DeepSeek-V3.2~\cite{liu2025deepseek}, with search tools to aggregate recently surfaced entities mentioned in high-coverage news or official websites w.r.t the sub-domain.
This process yields about 50K seed entities, denoted as $\mathrm{E}$. Illustration examples of seed entity collection are provided in Appendix~\ref{appendix:prompt_of_seed_entity}.

\paragraph{QA Synthesis Agent.}

For a given seed entity $e \in \mathrm{E}$, the QA synthesis agent iteratively mines information from the live web to construct a QA pair $\langle q, a \rangle$.
The overall synthesis process can be formulated as a $m$-step iterative chain:
\begin{equation}
\small
e \rightarrow \langle q_e^{(1)}, a_e^{(1)} \rangle \rightarrow \langle q_e^{(2)}, a_e^{(2)} \rangle \rightarrow \dots \rightarrow \langle q_e^{(m)}, a_e^{(m)} \rangle
\end{equation}
where $q_e^{(t)}, a_e^{(t)}$ denotes the question and its answer generated in $t$-th iteration, respectively. In detail, the agent involves two sub-steps in each iteration:

\vspace{0.5ex}
\noindent (1) \emph{Web Information Gathering}:
The agent collects information via engaging in multi-turn interactions with web tools: a \emph{search} tool uses the Google search engine to retrieve information and a \emph{visit} tool extracts targeted information from specific web pages.
In the course of this multi-turn interaction, we encourage the agent to gather \emph{fresh} knowledge, defined as information that becomes available after a specified timestamp $t$.\footnote{In this paper, we set $t$ to January 1, 2026, and it can be trivially adjusted to other timestamps (\emph{e.g.}, training cutoffs of specific LLMs).}
The agent then refines the gathered knowledge into an evidence list, denoted as $\mathcal{E} = \{\epsilon_1, \epsilon_2, ..., \epsilon_n\}$, where each $\epsilon_i$ indicates a concise knowledge statement (\emph{e.g.}, entity $e_i$ has some specific attributes).

\vspace{0.5ex}
\noindent (2) \emph{QA Pair Construction}:
Leveraging the evidence list $\mathcal{E}$, the agent incorporates these pieces of evidence to synthesize a complex QA pair $\langle q_e^{(t)}, a_e^{(t)} \rangle$.
Ideally, all evidence in $\mathcal{E}$ is fresh knowledge, ensuring that the synthesized $q_e^{(t)}$ is free from data contamination because it falls completely outside the search agents’ parametric memorization.
However, fresh knowledge appears much less frequently on the live web than its counterpart, namely non‑fresh knowledge (\emph{i.e.}, information already available before the timestamp $t$).
Consequently, although we encourage the agent to gather fresh knowledge, $\mathcal{E}$ inevitably contains non‑fresh knowledge.
If we strictly require all evidence to be fresh knowledge, the scale of $\mathcal{E}$ will be too limited to synthesize a complex question.
Therefore, we allow some non-fresh knowledge in $\mathcal{E}$, and require the agent to classify each $\epsilon_i$ as either fresh or non-fresh.
To avoid overly covered answers induced by the non‑fresh knowledge in $\mathcal{E}$, we limit the final answer to be based on fresh knowledge.
In this way, a preliminary question $\hat{q}_e^{(t)}$ together with its answer ${a}_e^{(t)}$ is generated.
To further enhance the difficulty of $\hat{q}_e^{(t)}$, wo follow \citet{li2025websailor,lu2025deepdive} and obfuscate features and relationships within $\hat{q}_e^{(t)}$ (\emph{e.g.}, vague time references and and non-specific descriptors) to obtain final question $q_e^{(t)}$.
The prompts employed by the QA synthesis agent in these two sub-steps are presented in the Appendix~\ref{appendix:prompt_of_qa_agent}.

\paragraph{Information Filtering Agent.}
Directly using the gathered evidence list $\mathcal{E}$ to synthesize QA pairs might involve the following issues:
(1) The evidence might suffer from rumor propagation, making the synthesized QA pairs unreliable.
This issue is particularly pronounced in the context of fresh knowledge compared to non-fresh knowledge~\cite{alkhodair2020detecting}.
(2) Too popular or over-covered non-fresh knowledge in $\mathcal{E}$ will make the relevant reasoning in the QA pairs too predictable for search agents~\cite{lu2025deepdive}.

To deal with the above issues, we introduce an information filtering agent to filter out flawed $\epsilon_i \in \mathcal{E}$:
(1) For each fresh $\epsilon_i$, the agent equipped with web tools (\emph{i.e.}, search and visit) to cross-validate the reliability of $\epsilon_i$ on the live web, and finally output a reliability label, \emph{i.e.}, ``credible'', ``not credible'', or ``unclear''.
(2) For each non-fresh $\epsilon_j$, the agent directly judges whether it is too popular or overly covered, without using any tools. It will output a popularity label, \emph{i.e.}, ``popular'' or ``non-popular''.
Only credible fresh evidence and non-popular non-fresh evidence are retained in $\mathcal{E}$.
If the length of $\mathcal{E}$ is less than a pre-defined threshold $k$, the QA synthesis agent retries the web information gathering process to collect additional evidence.
The prompts used to obtain the reliability and popularity labels are provided in the Appendix~\ref{appendix:prompt_of_reliability_agent}.

\begin{figure}[t]
\centerline{\includegraphics[width=0.45\textwidth]{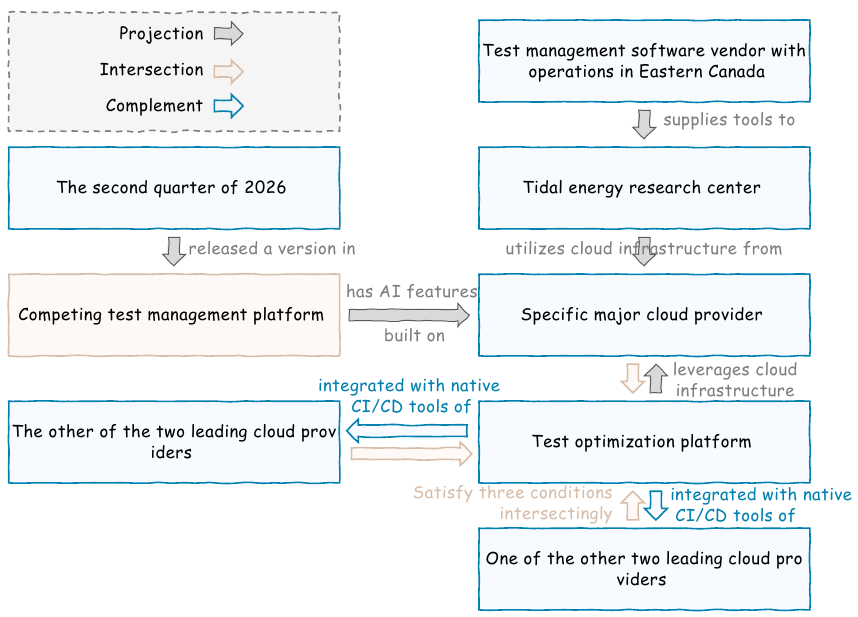}}
\caption{The illustration of a reasoning graph.}
\label{fig:reasoning_graph}
\end{figure} 

\paragraph{High-Level Guidance Agent.}
As pointed out by \citet{tao2025webshaper}, information-driven text-only synthesis paradigms struggle to capture underlying complex topologies (\emph{e.g.}, entity–relation structures) and lack systematic control.
As a result, the synthesized QA pairs tend to be \emph{redundant} or \emph{shortcut-prone} reasoning paths.
In contrast, graphs offer a structured, semantically rich environment for multi-hop reasoning, enabling explicit control over reasoning paths~\cite{lu2025deepdive}.

To mitigate reasoning redundancy and shortcuts inherent in text-only synthesis paradigms and to provide high-level, explicit control over QA synthesis, we introduce a high-level guidance agent.
This agent structures each synthesized question $q_e^{(t)}$ into a reasoning graph $\mathcal{G}_e^t = \{\mathcal{V}_e^t, \mathcal{R}_e^t\}$, where $\mathcal{V}_e^t$ and $\mathcal{R}_e^t$ denote the note set and the edge set in $\mathcal{G}_e^t$, respectively.
Figure~\ref{fig:reasoning_graph} shows an example of $\mathcal{G}_e^t$ using the example question in Figure~\ref{fig:example_question}.
Each node $v_i \in \mathcal{V}_e^t$ indicates an object, \emph{i.e.}, an entity (set) or an attribute (set).
As for edges $\mathcal{R}_e^t$, we employ logical operations to capture the complex topologies within $\mathcal{V}_e^t$:
(1) \textbf{Intersection} returns the intersection of $v_i$ and $v_j$, denoted as $v_i \cap v_j$.
For example, the intersection of Turing Award laureates and females is the set of female Turing Award laureates.
(2) \textbf{Complement} returns the complement of $v_i$, denoted as $\overline{v_i}$.
For example, the complement of Turing Award laureates (with respect to all people) is the set of people who have not received the Turing Award.
These two operations form a complete basis for expressing any entity set~\cite{enderton2001mathematical}.
To further support relations and attributes, we introduce: (3) \textbf{Projection}, which projects $v_i$ to $v_j$ via a specific relation $r$, denoted as $v_j = \pi_r(v_i)$. For example, $\pi_{\text{winner}}(\text{Turing Award})$ denotes Turing Award laureates, while $\pi_{\text{gender}}(\text{Jackie Chan})$ denotes Jackie Chan's gender.
Based on the above definition, each edge in the graph is denoted by one of the three operations.

After parsing $q_e^{(t)}$ into $\mathcal{G}_e^t$, we (1) detect reasoning redundancy by checking for isolated nodes or subgraphs in $\mathcal{G}_e^t$, and (2) detect reasoning shortcuts by examining whether there exist structural bypasses leading to the answer node. These detections can be achieved by an off-the-shelf toolkit, \emph{i.e.}, NetworkX\footnote{\url{https://github.com/networkx/networkx}}.
Further, using $\mathcal{G}_e^t$ together with the detected reasoning redundancies and shortcuts, the high-level guidance agent generates a textual instruction (denoted as $\mathcal{I}_e^t$) that specifies the synthesis direction (\emph{e.g.}, which logical operations to add), and any redundancy or shortcuts should be avoided in the subsequent iteration. The instructions are fed into the QA synthesis agent and guide its behavior in the next iteration:
\begin{equation}
\small
\big( \mathcal{I}_e^t, \langle q_e^{(t)}, a_e^{(t)} \rangle\big) \xrightarrow{\text{QA Synthesis Agent}} \langle q_e^{(t+1)}, a_e^{(t+1)} \rangle
\end{equation}

The prompts of graph parsing and instruction generation are provided in the Appendix~\ref{appendix:prompt_of_guidance_agent}.

\paragraph{Iteration Termination.}
The iteration of the three-agent collaborative framework is repeated until all of the following conditions are satisfied:
(1) the synthesized question $q_e^{(t)}$ contains no redundancy or shortcuts;
(2) at least five iterations have been executed;
(3) the reasoning graph contains at least five edges, \emph{i.e.}, $|\mathcal{R}_e^t|\geq5$.

\subsection{Data Quality}

In the previous section, we mainly use the high-level guidance agent to control the data quality, \emph{i.e.}, avoid reasoning redundancy and shortcuts in the synthesized questions.
To further ensure the textual quality, uniqueness and difficulty of the collected QA pairs, we employ the following strategies:

\paragraph{Textual Quality.}

We evaluate whether each synthesized QA pair is fluent, clear, and unambiguous by using DeepSeek-V3.2~\cite{deepseekv3.2} as the judge model. Then, we filter out the low-quality QA pairs. The prompt is shown in Appendix~\ref{appendix:qa_filtering}.

\paragraph{Uniqueness and Difficulty.}

To avoid alternative answers in the synthesized questions, we adopt a cross‑validation method inspired by xbench-DeepSearch~\cite{xbench}. In detail, for each question, we employ six cutting-edge LLMs\footnote{DeepSeek-V4, DeepSeek-V3.2, GLM-5, Kimi-K2.6, Qwen3.5-397B-A17B and Qwen3.5-122B-A10B.} as search agents to answer the question three times independently\footnote{Temperature is set to 1.0; top\_p is set to 0.95.}, resulting in 18 solutions.
If more than 80\% of the solutions converge on the same incorrect answer, the question is treated as a multiple-answer question and is discarded.
As for difficulty, if more than five (out of six) LLMs correctly answer the question, it will be discarded.

\begin{table}[t]
\centering
\resizebox{0.48\textwidth}{!}
{
\begin{tabular}{lcc}
\toprule[1pt]
 & \multicolumn{1}{c}{Gave up after 2h}  & \multicolumn{1}{c}{Solved within 2h}  \\ \midrule[1pt]
EvoBrowseComp-EN    & 76.0\% & 24.0\%       \\
EvoBrowseComp-ZH    & 80.0\% & 20.0\%       \\ \midrule[1pt]
Overall    & 78.0\% & 22.0\%       \\ \bottomrule[1pt]
\end{tabular}
}
\caption{Human baseline results on EvoBrowseComp under a 2h-per-question limit.}
\label{tab:human_baseline}
\end{table}

\paragraph{Human Validation.}

After quality filtering, we balance the data distribution for each domain and ultimately obtain 400 English and 400 Chinese QA pairs, which constitute our EvoBrowseComp.
To further reveal the data quality, we conduct human analyses on randomly selected 400 QA samples (200 English and 200 Chinese).
Since directly asking humans to answer questions in a web environment is overly challenging, we instead use the evidence list $\mathcal{E}$ as an anchor and require human evaluators to verify (1) whether each $\epsilon_i \in \mathcal{E}$ is correct, and (2) whether each synthesized question is consistent with its corresponding evidence list $\mathcal{E}$ and is unambiguous; (3) whether each answer can be inferred by the evidence list $\mathcal{E}$ (for more details, please refer to Appendix~\ref{appendix:data_quality}).
The results indicate that 94.2\% of evidence lists are entirely correct, and 90.8\% of questions are both consistent with their corresponding evidence lists and unambiguous.\footnote{The remaining 5.8\% of evidence lists may involve hallucinations.}
100.0\% of answers can be inferred from $\mathcal{E}$.
Overall, \emph{87.8\%} of QA pairs simultaneously pass the above three verifications, indicating the superiority of our synthesis framework.

Following BrowseComp~\cite{wei2025browsecomp}, we further conduct a human baseline to reveal the difficulty of EvoBrowseComp.
We recruit 6 evaluators and ask each to independently solve 100 QAs under a 2h-per-question limit, during which they may use a web browser but no AI assistant (details are provided in Appendix~\ref{appendix:human_baseline}).
As shown in Table~\ref{tab:human_baseline}, human evaluators solve only 22.0\% of the QAs and give up on 78.0\% after 2h.

\subsection{Data Statistics}

\begin{figure}[t]
\centerline{\includegraphics[width=0.45\textwidth]{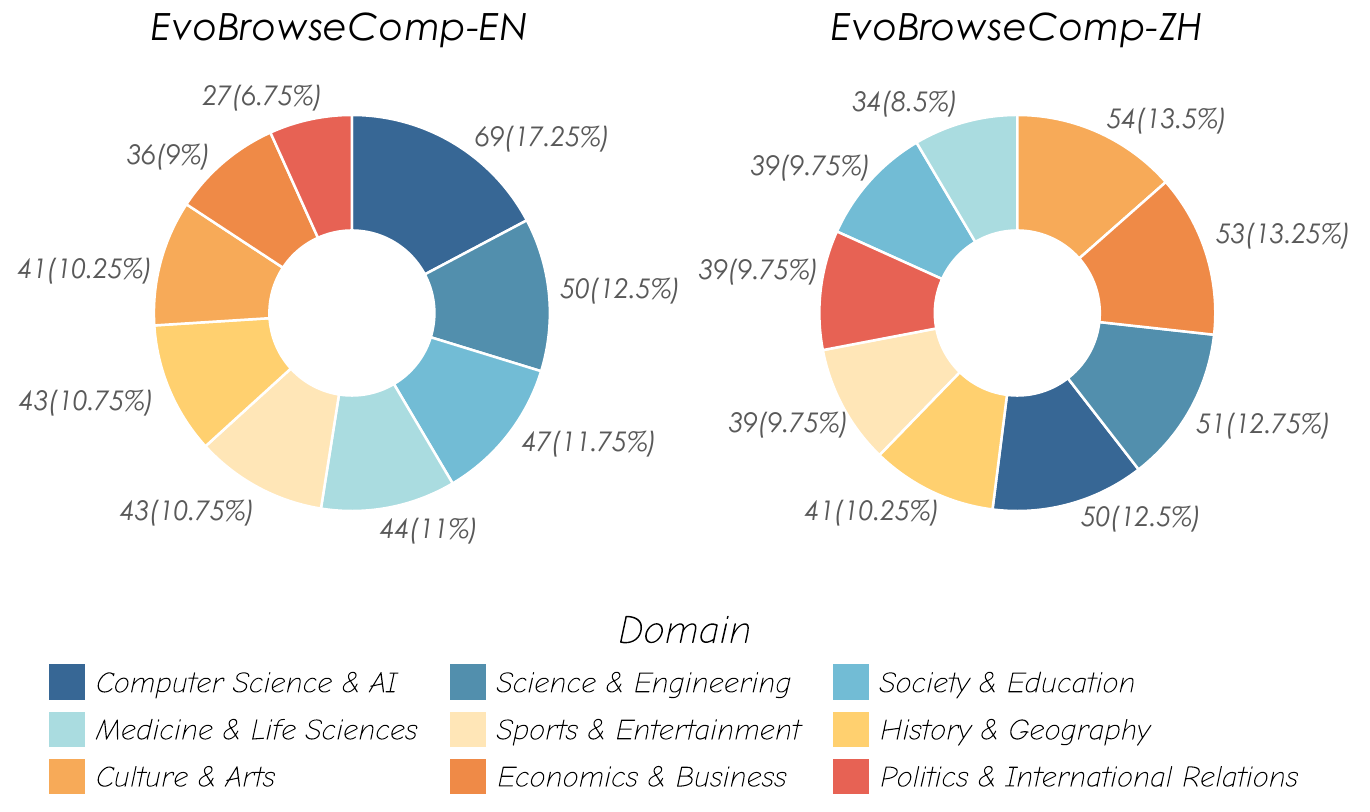}}
\caption{Distribution across nine domains in EvoBrowseComp.}
\label{fig:domian}
\end{figure}

\begin{table}[t]
\centering
\resizebox{0.48\textwidth}{!}
{
\begin{tabular}{lcccc}
\toprule[1pt]
\multicolumn{1}{c}{Data} & \multicolumn{1}{c}{Scale}  & \multicolumn{1}{c}{Lang.} &   \multicolumn{1}{c}{Length} & \multicolumn{1}{c}{Node}  \\ \midrule[1pt]
BrowseComp~\cite{wei2025browsecomp}      & 1266 &  En & 128.86  & 8.85       \\
DeepSearchQA~\cite{gupta2026deepsearchqa}     & 900 & En & 65.33  & 5.83       \\ 
EvoBrowseComp-EN (Our)    & 400 & En & 142.48    & 8.62     \\ \midrule[1pt]
BrowseComp-ZH~\cite{zhou2025browsecomp}    & 289 & Zh & 73.88   & 6.63      \\
WebWalkerQA~\cite{wu-etal-2025-webwalker}    & 680 & Zh & 33.99  & 4.63       \\
EvoBrowseComp-ZH (Our)    & 400 & Zh &  162.33  & 8.07   \\ \bottomrule[1pt]
\end{tabular}
}
\caption{Data Statistics of EvoBrowseComp compared with previous benchmark datasets (Lang.: Language). ``Length'' denotes the average question length, and ``Node'' denotes the average number of nodes in the reasoning graphs.}
\label{table:question_length}
\end{table}

EvoBrowseComp contains 800 high-quality QA pairs (400 in English and 400 in Chinese). We analyze the data statistics from the following aspects:

\vspace{0.5ex}
\noindent \textbf{Domain Distribution.}
As shown in Figure~\ref{fig:domian}, EvoBrowseComp is evenly distributed across nine predefined domains, ensuring broad coverage of the knowledge areas.

\vspace{0.5ex}
\noindent \textbf{Length.}
We calculate the average question length of EvoBrowseComp and previous benchmark datasets. As shown in Table~\ref{table:question_length},
The average question lengths in EvoBrowseComp are 142.48 and 162.33 tokens for English and Chinese, respectively, which are generally longer than those in previous datasets, particularly for Chinese.

\vspace{0.5ex}
\noindent \textbf{Complexity.}
We use reasoning graphs to structure complex questions in Section~\ref{subsec:2.1}.
In addition to EvoBrowseComp, we also extract reasoning graphs for complex questions of previous datasets.
The average number of nodes in reasoning graphs, which can reflect questions' complexity, is also reported in Table~\ref{table:question_length}:
The average number of nodes in EvoBrowComp-EN is similar to that in BrowseComp (8.62 vs. 8.85). On Chinese data, the average number of nodes in EvoBrowComp-Zh is significantly greater than others, \emph{i.e.}, 8.07 vs. 6.63/4.63, indicating the complexity of our data.

\begin{figure}[t]
\centerline{\includegraphics[width=0.45\textwidth]{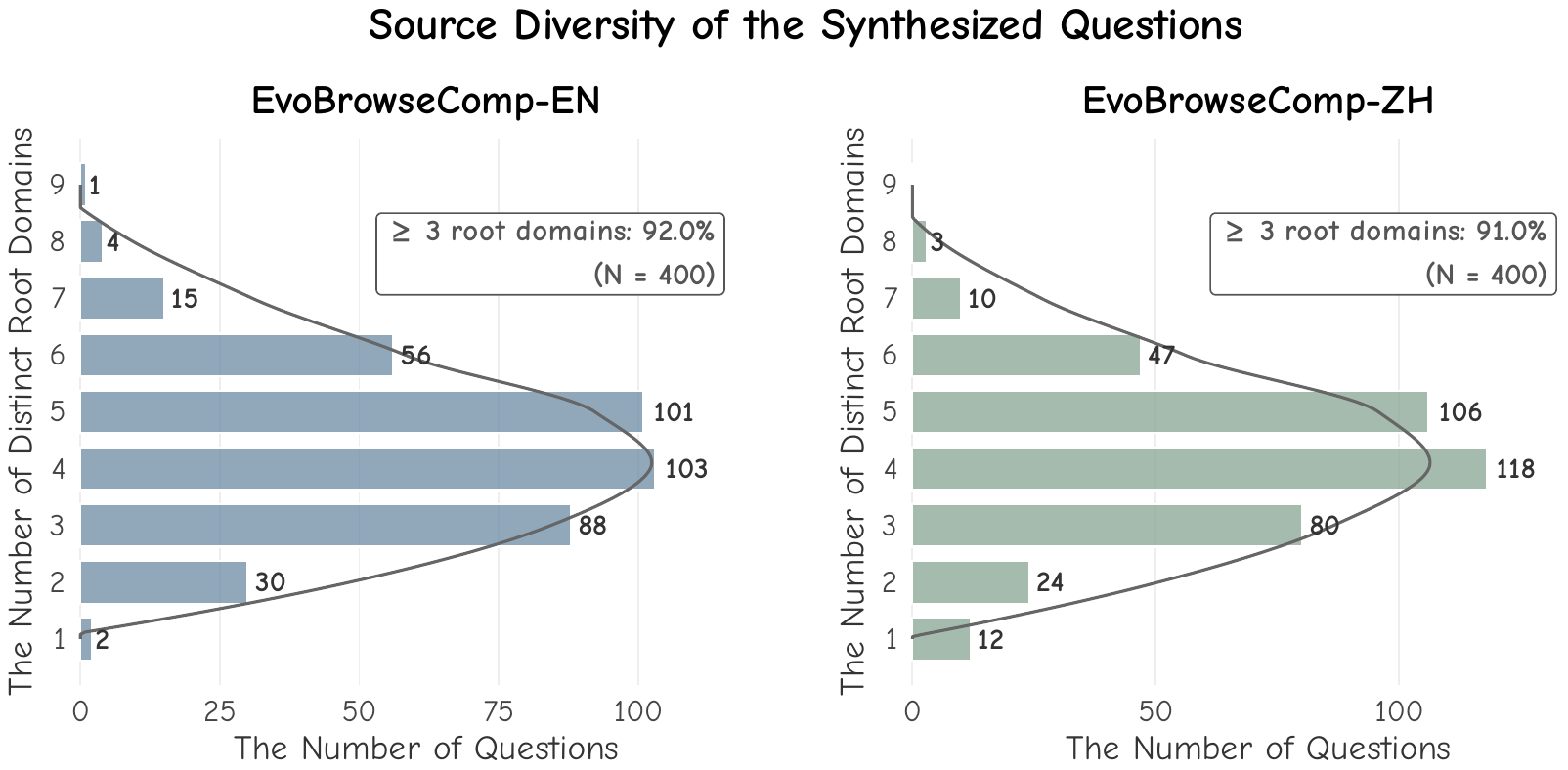}}
\caption{Distribution of the number of distinct root domains per question in EvoBrowseComp.}
\label{fig:url}
\end{figure} 

\vspace{0.5ex}
\noindent \textbf{Source Diversity.}
During collecting the evidence list $\mathcal{E}$ in data synthesis, we also record the source URL of each evidence $\epsilon_i \in \mathcal{E}$.
To characterize the source diversity of the synthesized questions, we calculate, for each question, the number of distinct root domains present in the corresponding $\mathcal{E}$.
For example, if the evidence for a given question is drawn from ``a.com'' and ``b.com'', there are two distinct root domains.
Theoretically, the greater the number of distinct root domains, the more complex and difficult the question becomes.
As shown in Figure~\ref{fig:url}, the number of distinct root domains per question in EvoBrowseComp exhibits a distinct bell-shaped distribution. On average, each question involves 4.2 distinct root domains, and over 90\% of questions require reasoning across at least three independent sources. This makes cross-site evidence aggregation and verification a necessary prerequisite for answering.

\subsection{Data Updation}

We intend to update the benchmark data annually for the following two reasons:
(1) Frequent updates are impractical due to the inherent scarcity of fresh knowledge (Section~\ref{subsec:2.1}).
(2) The high cost of data generation further renders more frequent updates unfeasible (Section~\ref{subsec:3.1}).

Each annual update consists of two major steps:
(1) Contamination Detection: We use the most advanced LLMs available at that time as search agents to answer each question without any web tools; questions that the majority can answer correctly are flagged as contaminated and removed.
(2) New Question Generation: By adjusting the timestamp $t$ and re-running the three-agent synthesis framework, we generate new complex questions from much fresher knowledge, keeping the benchmark up-to-date.

\subsection{Evaluation Protocol}

In model evaluation, search agents are required to answer the given complex questions under multi-turn interactions with the following two web tools:
(1) \emph{Search} uses the Google search engine to retrieve information.
(2) \emph{Visit} extracts targeted information from specific web pages.
The definition and implementation of these tools are derived from \citet{team2025tongyi}.\footnote{\url{https://github.com/Alibaba-NLP/DeepResearch}}
Following previous benchmark datasets~\cite{wei2025browsecomp,zhou2025browsecomp}, we also employ the LLM-as-a-judge paradigm in model evaluation.
Specifically, a judge model is used to verify whether a model prediction is correct or not.
The judge prompt is derived from \citet{team2025tongyi,xbench}, and is provided in the Appendix~\ref{appendix:judge_prompt}.
We use GLM-5-Chat~\cite{zeng2026glm} as the judge model since it shows a strong correlation with human judgments (please refer to the Appendix~\ref{appendix:judge_model}).
The evaluation metric is accuracy based on the model judgments.

\begin{table*}[t]
\centering
\resizebox{0.80\textwidth}{!}
{
\begin{tabular}{lcccc}
\toprule[1pt]
\multicolumn{1}{c}{\multirow{2}{*}{Model}}                       & \multicolumn{2}{c}{English} & \multicolumn{2}{c}{Chinese} \\
\multicolumn{1}{c}{}                       & w/ tools     & w/o tools    & w/ tools     & w/o tools    \\ \midrule[1pt]
Claude-Opus-4.8                            & \textbf{46.2}         & \textbf{9.0}            & \textbf{38.3}            & \textbf{14.0}            \\
Claude-Opus-4.6                            & \underline{44.8}         & 6.0            & \underline{36.8}            & 8.8            \\
GLM-5                                      & 39.2         & 1.1          & 30.5         & 2.5          \\
Kimi-K2.6                                  & 31.0         & 0.0          & 19.8         & 1.6          \\ 
Qwen3.5-397B-A17B-FP8                      & 42.0         & 2.9          & 34.5         & 3.1          \\
Qwen3.5-122B-A10B                          & 29.2         & 2.9          & 21.0         & 2.6          \\
Qwen3.5-35B-A3B                            & 29.2         & 0.8          & 14.2         & 1.2          \\
Qwen3.5-27B                                & 25.0         & 2.8          & 18.6         & 4.5          \\
Qwen3-235B-A22B-Thinking-2507              & 16.5         & 3.5          & 17.5         & 7.0          \\
DeepSeek-V4-Pro                            & 16.5         & 0.8          & 15.0         & 3.5          \\
DeepSeek-V4-Flash                          & 16.5         & 2.0          & 10.8         & 4.5          \\
DeepSeek-V3.2                              & 23.0         & \underline{6.3}          & 30.5         & \underline{10.3}        \\ \bottomrule[1pt]
\end{tabular}
}
\caption{Experimental results on EvoBrowseComp. The \textbf{bold} and the \underline{underline} denote the best and second-best performances, respectively.}
\label{table:main_results}
\end{table*}

\section{Experiments}

\subsection{Experimental Setup}
\label{subsec:3.1}

\noindent \textbf{LLMs.} Based on our EvoBrowseComp, we evaluate the following cutting-edge LLMs:
Claude-Opus-4.8~\cite{Claude4.8}, Claude-Opus-4.6~\cite{Claude4.6}, 
Qwen3.5-397B-A17B-FP8\footnote{\url{https://huggingface.co/Qwen/Qwen3.5-397B-A17B-FP8}}, Qwen3.5-122B-A10B\footnote{\url{https://huggingface.co/Qwen/Qwen3.5-122B-A10B}}, Qwen3.5-35B-A3B\footnote{\url{https://huggingface.co/Qwen/Qwen3.5-35B-A3B}}, Qwen3.5-27B\footnote{\url{https://huggingface.co/Qwen/Qwen3.5-27B}}~\cite{Qwen3.5}, Qwen3-235B-A22B-Thinking-2507\footnote{\url{https://huggingface.co/Qwen/Qwen3-235B-A22B-Thinking-2507}}~\cite{yang2025qwen3},
DeepSeek-V4-Pro\footnote{\url{https://huggingface.co/deepseek-ai/DeepSeek-V4-Pro}}, DeepSeek-V4-Flash\footnote{\url{https://huggingface.co/deepseek-ai/DeepSeek-V4-Flash}}~\cite{deepseekai2026deepseekv4}, DeepSeek-V3.2\footnote{\url{https://huggingface.co/deepseek-ai/DeepSeek-V3.2}}~\cite{deepseekv3.2}, 
GLM-5\footnote{\url{https://huggingface.co/zai-org/GLM-5}}~\cite{zeng2026glm} and Kimi-K2.6\footnote{\url{https://huggingface.co/moonshotai/Kimi-K2.6}}~\cite{kimik26}.

\vspace{0.5ex}
\noindent \textbf{Implementation Details.}
In data collection, all three agents are employed based on DeepSeek‑V3.2~\cite{deepseekv3.2}, which is self‑hosted (16 NVIDIA H20 GPUs per deployment). The pre‑defined threshold $k$ is set to 5 in the information filtering agent. The whole data construction takes $\sim40$K GPU hours, with $\sim57$K search and $\sim64$K visit tool calls.
In model evaluation, all open-source LLMs are deployed on NVIDIA H20 GPUs (96G) using SGLang\footnote{\url{https://github.com/sgl-project/sglang}}.
Most models are run on 8 GPUs, except DeepSeek-V4-Pro and GLM-5, which each use 32 GPUs, and DeepSeek-V3.2, which uses 16 GPUs.
We adopt sampling-based decoding with a temperature of 0.6 and a top\_p of 0.95.
We set the maximum context length to 128K for all LLMs to ensure a fair comparison, and limit the maximum number of tool calls to 40.
All LLMs are evaluated with the (maximum) thinking mode.
After model prediction, we use GLM-5-Chat (zero temperature) as the judge model.
For each LLM, we run three independent evaluations and report the average.

\begin{table}[t]
\centering
\resizebox{0.48\textwidth}{!}
{
\begin{tabular}{lcccc}
\toprule[1pt]
\multicolumn{1}{c}{Model} & \multicolumn{1}{c}{BC}  & \multicolumn{1}{c}{EvoBC.-EN} & \multicolumn{1}{c}{BC-ZH}  & \multicolumn{1}{c}{EvoBC.-ZH} \\ \midrule[1pt]
DeepSeek-V3.2      & 51.4 &  23.0 & 65.0 &   30.5      \\
GLM-5    & 62.0 & 39.2 & 72.7 &  30.5    \\ 
Qwen3.5-397B    & 69.0 & 42.0 &  70.3 & 34.5   \\ \bottomrule[1pt]
\end{tabular}
}
\caption{The performance of three example LLMs on BrowseComp (BC), BrowseComp-ZH (BC-ZH) and EvoBrowseComp-EN/ZH (EvoBC.-EN/ZH).}
\label{table:other_data}
\end{table}

\subsection{Results \& Analyses}

Table~\ref{table:main_results} presents the model performances on EvoBrowseComp, and we analyze the results from the following aspects:

\vspace{0.5ex}
\noindent \textbf{Tool-Free Setting.}
Without tools, most LLMs' accuracy is below 5\% (\emph{e.g.}, Kimi-K2.6 only achieves 0\% and 1.6\% in English and Chinese, respectively), and even the best-performing LLM (Claude-Opus-4.8) only achieves 9.0\% in English and 14.0\% in Chinese.
The limited tool-free performance suggests that EvoBrowseComp effectively prevents reliance on parameterized memory by introducing fresh knowledge in the complex questions.

\vspace{0.5ex}
\noindent \textbf{Tool-Based Setting.}
With access to web tools, in English, Claude-Opus-4.8 ranks first with 46.2\%, followed by Claude-Opus-4.6 (44.8\%), and Qwen3.5-397B (42.0\%). The results of the Chinese evaluation also show similar trends.
Taking Qwen3.5-397B, GLM-5 and DeepSeek-V3.2 as example LLMs, we also compare their performance on EvoBrowseComp and previous BrowseComp(-ZH)~\cite{wei2025browsecomp,zhou2025browsecomp}.\footnote{The results on BrowseComp and BrowseComp-Zh are directly borrowed from \citet{zeng2026glm,deepseekv3.2,Qwen3.5}.}
As shown in Table~\ref{table:other_data}, we find that the model performances on our data are significantly lower than those on BrowseComp / BrowseComp-ZH, indicating the difficulty of our complex questions.

\begin{table}[t]
\centering
\resizebox{0.40\textwidth}{!}
{
\begin{tabular}{lcccc}
\toprule[1pt]
\multicolumn{1}{c}{\multirow{2}{*}{Model}}                       & \multicolumn{2}{c}{English} & \multicolumn{2}{c}{Chinese} \\
\multicolumn{1}{c}{}                       & Acc.     & ER.    & Acc.     & ER.   \\ \midrule[1pt]
DS-V4-Chat      &  27.2 & 51.8 & 20.0   & 47.5  \\
DS-V4-High   & 34.5  & 38.8 & 24.8  &  54.5 \\ 
DS-V4-Max    & 16.5 & 75.5 & 10.8 & 82.5 \\ 
Claude-Opus-4.6    & 44.8 & 26.2 & 36.8 & 25.2 \\ \bottomrule[1pt]
\end{tabular}
}
\caption{The performance of DeepSeek-V4-Flash on EvoBrowseComp using different reasoning efforts. DS.: DeepSeek; ACC.: ``accuracy''; ER.: ``Exceed Ratio'' indicates the proportion of evaluation samples that exceed the maximum allowed number of tool calls.}
\label{table:dsv4_performance}
\end{table}

\vspace{0.5ex}

\noindent \textbf{Scaling Laws in Different Tool Settings.}
We also evaluate the performance of the Qwen3.5 family (27B → 35B-A3B → 122B-A10B → 397B-A17B), and identify the following scaling behaviors:
When tools are utilized, accuracy improves with increased model capability (EN: 25.0 → 42.0; ZH: 18.6 → 34.5 overall). In contrast, without tools, accuracy remains consistently low, hovering around 1–5\%, regardless of model scale. This indicates that no scaling law is evident, as the necessary episodic or fresh knowledge is not embedded within the parameters at any model size.

\noindent \textbf{The Effect of Reasoning Effort.}
We also observe that DeepSeek-V4-Pro/Flash underperforms DeepSeek-V3.2. We manually inspect the predictions of DeepSeek-V4 and find that many evaluation samples exceed the maximum allowed number of tool calls (40; c.f., \S~\ref{subsec:3.1}), which leads to its unsatisfactory performance.
To figure out the effect of reasoning effort, we use DeepSeek-V4-Flash as an example and evaluate its performance under three settings: the original max-reasoning setting, a high-reasoning setting (DS-V4-High), and a non-reasoning setting (DS-V4-Chat).
In addition to accuracy, we also report the proportion of evaluation samples that exceed the maximum allowed number of tool calls (abbr. ER).
As shown in Table~\ref{table:dsv4_performance}, we find that DeepSeek-V4-Flash-High achieves the best performance among the three settings, while DeepSeek-V4-Flash-Max even underperforms DeepSeek-V4-Flash-Chat.
In terms of ER, we find that DeepSeek-V4 achieves much high ER scores than Claude-Opus-4.6. For example, DeepSeek-V4-Max achieves 75.5\% and 82.5\% ER scores in English and Chinese.
This phenomenon raises concerns about reasoning efficiency. Although state-of-the-art LLMs exhibit strong reasoning capabilities, enhancing their reasoning efficiency remains crucial for practical applications.

\section{Related Work}

A growing body of work introduces benchmarks to evaluate the browsing, reasoning, and retrieval capabilities of LLMs. Early datasets such as NaturalQuestions~\cite{kwiatkowski-etal-2019-natural}, TriviaQA~\cite{joshi-etal-2017-triviaqa}, and HotpotQA~\cite{yang-etal-2018-hotpotqa} focus on single-hop or multi-hop fact retrieval; however, many of these datasets are effectively handled by cutting-edge LLMs~\cite{mialon2024gaia}.
To raise the difficulty ceiling, BrowseComp~\cite{wei2025browsecomp} introduces reverse-engineered questions requiring persistence and creativity in information seeking, while BrowseComp-Plus~\cite{chen2025browsecomp} provides a fixed, human-verified corpus to disentangle retriever performance from search agent reasoning. Parallel efforts such as BrowseComp-ZH~\cite{zhou2025browsecomp} extend BrowseComp to Chinese, and WebWalkerQA~\cite{wu-etal-2025-webwalker} emphasizes vertical web traversal through structured official websites. 
GAIA~\cite{mialon2024gaia} proposes conceptually simple yet execution-heavy tasks for general AI assistants, requiring diverse tool use and multi-step planning. DeepSearchQA~\cite{gupta2026deepsearchqa} shifts the evaluation focus from single-answer retrieval to exhaustive set generation, stressing systematic collation, entity resolution, and stopping criteria. More recently, SealQA~\cite{pham2025sealqa} highlights the brittleness of search-augmented models under conflicting, noisy, or misleading retrieval results. Despite their respective strengths, these benchmarks largely rely on \textit{static} or \textit{fixed} corpora. As noted by \citet{pham2025sealqa}, this static nature renders them susceptible to progressive data contamination.
Different from previous datasets, we design a three-agent framework to automatically discover \emph{fresh} knowledge and construct contamination-free complex questions.
Without costly manual annotation, our data can be regularly updated to prevent data contamination and ensure temporal freshness.

\section{Conclusion}

In this paper, we introduce EvoBrowseComp, an evolving search agent benchmark, which contains 400 English and 400 Chinese contamination-free complex QA pairs.
We design a three-agent collaborative framework to discover fresh knowledge from the live web and synthesize the QA data.
Multiple strategies are implemented to ensure data quality, including the avoidance of reasoning redundancy and shortcuts, as well as the verification of textual quality, answer uniqueness, and question difficulty.
Human analyses further indicate that the synthesized data achieves a high level of quality.
Since the framework operates entirely automatically and does not require costly manual annotation, it can be regularly updated to prevent data contamination and ensure temporal freshness.
Furthermore, experimental results on cutting-edge LLMs underscore the challenges posed by this data. Thus, it establishes a sustainable paradigm for the future-proof evaluation of search agents.

\section*{Limitations}

While we show the effectiveness of the three-agent synthesis framework, there are some limitations worth noting:
(1) We employ DeepSeek-V3.2~\cite{deepseekv3.2} as the backbone of the three agents, and thus, the synthesized data might involve the same biases and toxic behaviors exhibited by the model.
(2) In the model evaluation, we use the judge model to assess only the final answers rather than the entire reasoning trajectory. Consequently, it becomes difficult to distinguish an agent that reasoned correctly from one that obtained the correct answer through inefficient or accidental means (\emph{e.g.}, lucky guessing).

\section*{Ethical Considerations}

We discuss the main ethical considerations of EvoBrowseComp as follows:
(1) Licenses. We will release our synthesized data under CC-BY-NC-SA 4.0 license.
(2) Privacy Information. We extract knowledge from the publicly available web pages, and we filter out potential privacy information via LLMs.

\bibliography{custom}

@article{chen2025browsecomp,
  title={Browsecomp-plus: A more fair and transparent evaluation benchmark of deep-research agent},
  author={Chen, Zijian and Ma, Xueguang and Zhuang, Shengyao and Nie, Ping and Zou, Kai and Liu, Andrew and Green, Joshua and Patel, Kshama and Meng, Ruoxi and Su, Mingyi and others},
  journal={arXiv preprint arXiv:2508.06600},
  year={2025}
}

@article{zhou2025browsecomp,
  title={Browsecomp-zh: Benchmarking web browsing ability of large language models in chinese},
  author={Zhou, Peilin and Leon, Bruce and Ying, Xiang and Zhang, Can and Shao, Yifan and Ye, Qichen and Chong, Dading and Jin, Zhiling and Xie, Chenxuan and Cao, Meng and others},
  journal={arXiv preprint arXiv:2504.19314},
  year={2025}
}

@article{wei2025browsecomp,
  title={Browsecomp: A simple yet challenging benchmark for browsing agents},
  author={Wei, Jason and Sun, Zhiqing and Papay, Spencer and McKinney, Scott and Han, Jeffrey and Fulford, Isa and Chung, Hyung Won and Passos, Alex Tachard and Fedus, William and Glaese, Amelia},
  journal={arXiv preprint arXiv:2504.12516},
  year={2025}
}

@article{gupta2026deepsearchqa,
  title={DeepSearchQA: Bridging the Comprehensiveness Gap for Deep Research Agents},
  author={Gupta, Nikita and Chatterjee, Riju and Haas, Lukas and Tao, Connie and Wang, Andrew and Liu, Chang and Oiwa, Hidekazu and Gribovskaya, Elena and Ackermann, Jan and Blitzer, John and others},
  journal={arXiv preprint arXiv:2601.20975},
  year={2026}
}

@inproceedings{
mialon2024gaia,
title={{GAIA}: a benchmark for General {AI} Assistants},
author={Gr{\'e}goire Mialon and Cl{\'e}mentine Fourrier and Thomas Wolf and Yann LeCun and Thomas Scialom},
booktitle={The Twelfth International Conference on Learning Representations},
year={2024},
url={https://openreview.net/forum?id=fibxvahvs3}
}

@article{pham2025sealqa,
  title={SealQA: Raising the Bar for Reasoning in Search-Augmented Language Models},
  author={Pham, Thinh and Nguyen, Nguyen and Zunjare, Pratibha and Chen, Weiyuan and Tseng, Yu-Min and Vu, Tu},
  journal={arXiv preprint arXiv:2506.01062},
  year={2025}
}

@inproceedings{wu-etal-2025-webwalker,
    title = "{W}eb{W}alker: Benchmarking {LLM}s in Web Traversal",
    author = "Wu, Jialong  and
      Yin, Wenbiao  and
      Jiang, Yong  and
      Wang, Zhenglin  and
      Xi, Zekun  and
      Fang, Runnan  and
      Zhang, Linhai  and
      He, Yulan  and
      Zhou, Deyu  and
      Xie, Pengjun  and
      Huang, Fei",
    editor = "Che, Wanxiang  and
      Nabende, Joyce  and
      Shutova, Ekaterina  and
      Pilehvar, Mohammad Taher",
    booktitle = "Proceedings of the 63rd Annual Meeting of the Association for Computational Linguistics (Volume 1: Long Papers)",
    month = jul,
    year = "2025",
    address = "Vienna, Austria",
    publisher = "Association for Computational Linguistics",
    url = "https://aclanthology.org/2025.acl-long.508/",
    doi = "10.18653/v1/2025.acl-long.508",
    pages = "10290--10305",
    ISBN = "979-8-89176-251-0"
}

@inproceedings{
patil2025the,
title={The Berkeley Function Calling Leaderboard ({BFCL}): From Tool Use to Agentic Evaluation of Large Language Models},
author={Shishir G Patil and Huanzhi Mao and Fanjia Yan and Charlie Cheng-Jie Ji and Vishnu Suresh and Ion Stoica and Joseph E. Gonzalez},
booktitle={Forty-second International Conference on Machine Learning},
year={2025},
url={https://openreview.net/forum?id=2GmDdhBdDk}
}

@article{liu2025deepseek,
  title={Deepseek-v3. 2: Pushing the frontier of open large language models},
  author={Liu, Aixin and Mei, Aoxue and Lin, Bangcai and Xue, Bing and Wang, Bingxuan and Xu, Bingzheng and Wu, Bochao and Zhang, Bowei and Lin, Chaofan and Dong, Chen and others},
  journal={arXiv preprint arXiv:2512.02556},
  year={2025}
}

@online{Anthropic2026EvalAwareness,
  author    = {Anthropic},
  title     = {Eval awareness in Claude Opus 4.6's BrowseComp performance},
  year      = {2026},
  month     = {3},
  day       = {6},
  url       = {https://www.anthropic.com/engineering/eval-awareness-browsecomp}
}

@article{li2025websailor,
  title={Websailor: Navigating super-human reasoning for web agent},
  author={Li, K and Zhang, Z and Yin, H and others},
  journal={arXiv preprint arXiv:2507.02592},
  year={2025}
}

@article{lu2025deepdive,
  title={Deepdive: Advancing deep search agents with knowledge graphs and multi-turn rl},
  author={Lu, Rui and Hou, Zhenyu and Wang, Zihan and Zhang, Hanchen and Liu, Xiao and Li, Yujiang and Feng, Shi and Tang, Jie and Dong, Yuxiao},
  journal={arXiv preprint arXiv:2509.10446},
  year={2025}
}

@misc{Claude4.8, 
    title={System Card: Claude Opus 4.8}, 
    url={https://www-cdn.anthropic.com/0b4915911bb0d19eca5b5ee635c80fef830a37ea.pdf}, 
    author={Anthropic}, 
    year={2026}
}

@misc{Claude4.6, 
    title={System Card: Claude Opus 4.6}, 
    url={https://www-cdn.anthropic.com/0dd865075ad3132672ee0ab40b05a53f14cf5288.pdf}, 
    author={Anthropic}, 
    year={2026}
}

@misc{deepseekai2026deepseekv4,
      title={DeepSeek-V4: Towards Highly Efficient Million-Token Context Intelligence},
      author={DeepSeek-AI},
      year={2026},
}

@misc{kimik26,
      title={Kimi K2.6: Advancing Open-Source Coding},
      author={Kimi},
      year={2026},
      url={https://www.kimi.com/blog/kimi-k2-6}
}

@article{deepseekv3.2,
  author       = {DeepSeek{-}AI},
  title        = {DeepSeek-V3.2: Pushing the Frontier of Open Large Language Models},
  journal      = {CoRR},
  volume       = {abs/2512.02556},
  year         = {2025},
  url          = {https://doi.org/10.48550/arXiv.2512.02556},
  doi          = {10.48550/ARXIV.2512.02556},
  eprinttype    = {arXiv},
  eprint       = {2512.02556},
  bibsource    = {dblp computer science bibliography, https://dblp.org}
}

@misc{Qwen3.5,
    title={Qwen3.5: Towards Native Multimodal Agents},
    url={https://qwen.ai/blog?id=qwen3.5},
    author={Qwen},
    year={2026}
}

@article{zeng2026glm,
  title={Glm-5: from vibe coding to agentic engineering},
  author={Zeng, Aohan and Lv, Xin and Hou, Zhenyu and Du, Zhengxiao and Zheng, Qinkai and Chen, Bin and Yin, Da and Ge, Chendi and Huang, Chenghua and Xie, Chengxing and others},
  journal={arXiv preprint arXiv:2602.15763},
  year={2026}
}

@article{yang2025qwen3,
  title={Qwen3 technical report},
  author={Yang, An and Li, Anfeng and Yang, Baosong and Zhang, Beichen and Hui, Binyuan and Zheng, Bo and Yu, Bowen and Gao, Chang and Huang, Chengen and Lv, Chenxu and others},
  journal={arXiv preprint arXiv:2505.09388},
  year={2025}
}

@article{tao2025webshaper,
  title={Webshaper: Agentically Data Synthesizing via Information-Seeking Formalization},
  author={Tao, Z and Wu, J and Yin, W and others},
  journal={arXiv preprint arXiv:2507.15061},
  year={2025}
}

@misc{xbench,
  title = {Xbench-DeepSearch},
  author = {Xbench-Team},
  url = {https://xbench.org/agi/aisearch},
  year={2025}
}

@article{team2025tongyi,
  title={Tongyi deepresearch technical report},
  author={Team, Tongyi DeepResearch and Li, Baixuan and Zhang, Bo and Zhang, Dingchu and Huang, Fei and Li, Guangyu and Chen, Guoxin and Yin, Huifeng and Wu, Jialong and Zhou, Jingren and others},
  journal={arXiv preprint arXiv:2510.24701},
  year={2025}
}

@article{alkhodair2020detecting,
  title={Detecting breaking news rumors of emerging topics in social media},
  author={Alkhodair, Sarah A and Ding, Steven HH and Fung, Benjamin CM and Liu, Junqiang},
  journal={Information Processing \& Management},
  volume={57},
  number={2},
  pages={102018},
  year={2020},
  publisher={Elsevier}
}

@book{enderton2001mathematical,
  title={A Mathematical Introduction to Logic},
  author={Enderton, H B},
  publisher={Elsevier},
  year={2001}
}

@article{kwiatkowski-etal-2019-natural,
    title = "Natural Questions: A Benchmark for Question Answering Research",
    author = "Kwiatkowski, Tom  and
      Palomaki, Jennimaria  and
      Redfield, Olivia  and
      Collins, Michael  and
      Parikh, Ankur  and
      Alberti, Chris  and
      Epstein, Danielle  and
      Polosukhin, Illia  and
      Devlin, Jacob  and
      Lee, Kenton  and
      Toutanova, Kristina  and
      Jones, Llion  and
      Kelcey, Matthew  and
      Chang, Ming-Wei  and
      Dai, Andrew M.  and
      Uszkoreit, Jakob  and
      Le, Quoc  and
      Petrov, Slav",
    editor = "Lee, Lillian  and
      Johnson, Mark  and
      Roark, Brian  and
      Nenkova, Ani",
    journal = "Transactions of the Association for Computational Linguistics",
    volume = "7",
    year = "2019",
    address = "Cambridge, MA",
    publisher = "MIT Press",
    url = "https://aclanthology.org/Q19-1026/",
    doi = "10.1162/tacl_a_00276",
    pages = "452--466"
}

@inproceedings{joshi-etal-2017-triviaqa,
    title = "{T}rivia{QA}: A Large Scale Distantly Supervised Challenge Dataset for Reading Comprehension",
    author = "Joshi, Mandar  and
      Choi, Eunsol  and
      Weld, Daniel  and
      Zettlemoyer, Luke",
    editor = "Barzilay, Regina  and
      Kan, Min-Yen",
    booktitle = "Proceedings of the 55th Annual Meeting of the Association for Computational Linguistics (Volume 1: Long Papers)",
    month = jul,
    year = "2017",
    address = "Vancouver, Canada",
    publisher = "Association for Computational Linguistics",
    url = "https://aclanthology.org/P17-1147/",
    doi = "10.18653/v1/P17-1147",
    pages = "1601--1611"
}

@inproceedings{yang-etal-2018-hotpotqa,
    title = "{H}otpot{QA}: A Dataset for Diverse, Explainable Multi-hop Question Answering",
    author = "Yang, Zhilin  and
      Qi, Peng  and
      Zhang, Saizheng  and
      Bengio, Yoshua  and
      Cohen, William  and
      Salakhutdinov, Ruslan  and
      Manning, Christopher D.",
    editor = "Riloff, Ellen  and
      Chiang, David  and
      Hockenmaier, Julia  and
      Tsujii, Jun{'}ichi",
    booktitle = "Proceedings of the 2018 Conference on Empirical Methods in Natural Language Processing",
    month = oct # "-" # nov,
    year = "2018",
    address = "Brussels, Belgium",
    publisher = "Association for Computational Linguistics",
    url = "https://aclanthology.org/D18-1259/",
    doi = "10.18653/v1/D18-1259",
    pages = "2369--2380"
}

@article{zar2005spearman,
  title={Spearman rank correlation},
  author={Zar, Jerrold H},
  journal={Encyclopedia of biostatistics},
  volume={7},
  year={2005},
  publisher={Wiley Online Library}
}

@article{o3,
  author      = {{OpenAI Team}},
  title       = {{OpenAI o3 and o4-mini} System Card},
  year        = {2025},
  url         = {https://cdn.openai.com/o3-mini-system-card-feb10.pdf}
}

\appendix

\section{Construction Details}

\subsection{Examples of Seed Entity Construction}
\label{appendix:prompt_of_seed_entity}

\begin{promptbox}[title=Prompt for Seed Entity Construction]
\small

You are an expert in seed entity generation. Given a ``primary domain : sub-domain'' pair, your task is to aggregate entities that have recently surfaced (in 2026) under the specified sub-domain, as reported by authoritative news outlets or official websites.

\textbf{Input}
\begin{itemize}\itemsep0pt
    \item Primary domain: \texttt{\{primary\_domain\}}
    \item Sub-domain: \texttt{\{secondary\_domain\}}
    \item Number of seeds to generate: \texttt{\{num\_seeds\}}
\end{itemize}

\textbf{Tool}
\begin{itemize}\itemsep0pt
    \item \texttt{search(query)}: performs a live web search and returns a list of webpage snippets.
\end{itemize}

\textbf{Requirements}
\begin{enumerate}\itemsep0pt
    \item \textit{Freshness}: Each entity must have either newly emerged in 2026 or undergone a significant attribute change in 2026 (e.g., newly founded organizations, newly released products, newly appointed individuals, newly occurred events).
    \item \textit{Source reliability}: Each entity must be reported by authoritative news sources (e.g., Reuters, BBC, Xinhua) or official websites (e.g., .gov sites, official homepages, reputable encyclopedias). Avoid personal blogs, self-media, and content farms.
    \item \textit{Diversity}: Cover a broad range of entity types, including people, organizations, events, products, concepts, and locations.
    \item \textit{Multi-round search}: Issue multiple \texttt{search()} calls with diversified queries to comprehensively cover the different facets of the sub-domain.
\end{enumerate}

\textbf{Output Format}
\begin{verbatim}
{
  "domain": {
    "primary": "<primary domain>",
    "secondary": "<sub-domain>"},
  "total_count": {num_seeds},
  "seeds": [
    {"id": 1,
     "entity_name": "<entity name>",
     "is_fresh": true}
  ]
}
\end{verbatim}

Now begin: first plan your search queries, then perform multiple rounds of \texttt{search()}, and finally output \texttt{\{num\_seeds\}} seed entities.

\end{promptbox}

\subsection{Examples of the QA Synthesis Agent}
\label{appendix:prompt_of_qa_agent}

\begin{promptbox}[title=Prompt for Evidence Collection]
\small

\textbf{Role: Knowledge Collection Agent} \\
Starting from the given seed, perform multiple rounds of \texttt{search} and \texttt{visit} to collect structured evidence that will support downstream QA construction.

\textbf{Definition of Fresh Knowledge} \\
Events, relations, attributes updates that occurred on or after January 1, 2026.

\textbf{Tools}
\begin{itemize}\itemsep0pt
    \item \texttt{search(query)}: live web search.
    \item \texttt{visit(url)}: fetch the full content of a specific page.
\end{itemize}

\textbf{Strategy}
\begin{enumerate}\itemsep0pt
    \item \textit{Iterative exploration}: Starting from the seed, iteratively invoke search/visit, following hyperlinks to expand the entity network.
    \item \textit{Prioritize fresh knowledge}: Augment queries with time-aware keywords such as ``2026'', ``latest'', or ``current''.
    \item \textit{Mandatory verification}: Every piece of evidence must come from a page you have actually visited; do not record evidence based on search snippets alone.
    \item \textit{Chainability}: Evidence items should share entities with one another so that they can later be linked into evidence chains.
\end{enumerate}

\textbf{Output Format}

\begin{lstlisting}[
    basicstyle=\ttfamily\footnotesize,
    breaklines=true,
    breakatwhitespace=false,
    columns=fullflexible,
    keepspaces=true,
    frame=none,
    postbreak=\mbox{\textcolor{gray}{$\hookrightarrow$}\space},
]
[
  {
    "evidence_id": "e_001",
    "head_entity": "Entity A",
    "relation": "relation or attribute description",
    "tail_entity": "Entity B or attribute value",
    "evidence_text": "verbatim snippet from the source page",
    "source_url": "URL actually visited",
    "is_fresh_knowledge": true
  }
]
\end{lstlisting}

\textbf{Quantity Requirement} \\
Collect at least \textbf{\{num\_evidence\}} evidence items, with no less than 50\% marked as fresh knowledge.

\vspace{2pt}
\texttt{<seed>\{seed\}</seed>}

\end{promptbox}

\begin{promptbox}[title=Prompt for QA Construction]
\small

\textbf{Role: QA Construction Agent} \\
Based on the provided evidence list, construct \textbf{\{num\}} reasoning QA pairs.

\vspace{2pt}
\texttt{<evidence\_list>
\{evidence\_list\}
</evidence\_list>}

\textbf{Hard Requirements}

\textbf{1. Reasoning Chain}
\begin{itemize}\itemsep0pt
    \item \textit{At least 5 hops}, with each hop introducing a fresh entity or relation (no looping in place).
    \item \textit{The final hop must be fresh knowledge.} 
    
    (\texttt{is\_fresh\_knowledge = true})
\end{itemize}

\textbf{2. Answer} \\
The answer must fall into one of the following types to ensure uniqueness and evaluability:
\begin{itemize}\itemsep0pt
    \item Deterministic attributes (time, number, or proper nouns with unique referents).
    \item The intersection, difference, or ranking result derived from multiple entities along the chain.
\end{itemize}

\textbf{3. Obfuscation (Core)} \\
\textit{Goal}: prevent the model from identifying entities via memorization or a single search; the full reasoning chain must be traversed.
\begin{itemize}\itemsep0pt
    \item \textit{Non-fresh knowledge}: use descriptions whose individual words are common but whose combination points to a rare attribute.
    \item \textit{Fresh knowledge}: anchor the description on a non-core but retrievable attribute, and phrase it with moderate vagueness.
    \item \textit{Time / location}: replace explicit values with event anchors or rare attribute combinations.
    \item \textit{The answer entity must not be directly or indirectly hinted at within the question.}
\end{itemize}

\textbf{4. Anti-Shortcut} \\
The final interrogative clause must be tightly bound to the last unresolved entity in the reasoning chain, and must not pivot to a public event or globally known attribute.

\textbf{Output Format}

\begin{lstlisting}[
    basicstyle=\ttfamily\footnotesize,
    breaklines=true,
    breakatwhitespace=false,
    columns=fullflexible,
    keepspaces=true,
    frame=none,
    postbreak=\mbox{\textcolor{gray}{$\hookrightarrow$}\space},
]
[
  {
    "id": "qa_001",
    "question": "obfuscated question",
    "answer": "final answer",
    "hop_count": 5,
    "knowledge_order": "nonfresh-nonfresh-fresh-fresh-fresh",
    "reasoning_chain": [
      {
        "hop": 1, 
        "evidence_id": "e_xxx", 
        "entity": "...", 
        "relation": "...", 
        "is_fresh_knowledge": false
      }
    ]
  }
]
\end{lstlisting}

Output the JSON array directly.

\end{promptbox}

\subsection{Examples of Information Filtering Agent}
\label{appendix:prompt_of_reliability_agent}

\begin{promptbox}[title=Prompt for Fresh Knowledge Reliability Assessment]
\small

\textbf{Role: Knowledge Reliability Verification Assistant} \\
Use the \texttt{search} and \texttt{visit} tools to \textbf{cross-validate} the given evidence list and assess its reliability.

\textbf{Evidence List to Verify} \\
Each item is a triple \texttt{(head\_entity, relation, tail\_entity)} supported by \texttt{evidence\_text} from \texttt{source\_url}, and flagged as fresh knowledge via \texttt{is\_fresh\_knowledge}.

\vspace{2pt}
\texttt{\{evidence\_list\}}

\textbf{Tools}
\begin{itemize}\itemsep0pt
    \item \texttt{search(query)}: live web search.
    \item \texttt{visit(url)}: fetch the full content of a specific page.
\end{itemize}

\textbf{Verification Procedure}
\begin{enumerate}\itemsep0pt
    \item For each evidence item, extract the key fact represented by the triple \texttt{(head\_entity, relation, tail\_entity)}.
    \item Issue \texttt{search} queries to retrieve relevant results, and \texttt{visit} authoritative sources (official websites, mainstream media, reputable encyclopedias) when further confirmation is needed. The provided \texttt{source\_url} may be visited as one reference, but \textbf{must not} be counted as an independent corroboration.
    \item \textit{Cross-validation}: for each triple, find at least two independent sources that corroborate it.
    \item Assign an overall reliability label for the entire evidence list:
    \begin{itemize}\itemsep0pt
        \item \texttt{credible}: every triple is consistently supported by multiple independent and trustworthy sources.
        \item \texttt{not credible}: at least one triple is contradicted by sources, or supported only by untrustworthy sources.
        \item \texttt{unclear}: available information is insufficient to make a definitive judgment on one or more triples.
    \end{itemize}
\end{enumerate}

\textbf{Output Format} \\
Output only the following JSON. Do not add any additional text or code-block markers:

\begin{lstlisting}[
    basicstyle=\ttfamily\footnotesize,
    breaklines=true,
    breakatwhitespace=false,
    columns=fullflexible,
    keepspaces=true,
    frame=none,
    postbreak=\mbox{\textcolor{gray}{$\hookrightarrow$}\space},
]
{
  "reliability": "credible | not credible | unclear"
}
\end{lstlisting}

\end{promptbox}
\begin{promptbox}[title=Prompt for Non-fresh Knowledge Popularity Assessment]
\small

\textbf{Role: Knowledge Assessment Assistant} \\
Based \textbf{solely on your internal knowledge}, judge whether the given evidence list is too popular or overly covered.

\textbf{Evidence List to Assess} \\
Each item is a triple \texttt{(head\_entity, relation, tail\_entity)} supported by \texttt{evidence\_text} from \texttt{source\_url}, and flagged as non-fresh knowledge via \texttt{is\_fresh\_knowledge}.

\vspace{2pt}
\texttt{\{evidence\_list\}}

\textbf{Criteria}
\begin{itemize}\itemsep0pt
    \item \texttt{popular}: the entities or facts described by the triples have prominent, widely known features, allowing you to identify the referents directly and unambiguously.
    \item \texttt{non-popular}: the descriptions are relatively obscure or lack distinctive features, requiring additional clues to identify the referents.
\end{itemize}

\textbf{Instructions}
\begin{itemize}\itemsep0pt
    \item Assess every triple in the list, then assign one overall label:
    \begin{itemize}\itemsep0pt
        \item Label as \texttt{popular} only if the majority of triples clearly describe widely known entities or facts.
        \item Otherwise, label as \texttt{non-popular}.
    \end{itemize}
    \item If the content lies beyond your knowledge scope (e.g., involves future information or entirely unfamiliar concepts), label it as \texttt{non-popular}.
\end{itemize}

\textbf{Output Format} \\
Output only the following JSON. Do not add any additional text or code-block markers:

\begin{lstlisting}[
    basicstyle=\ttfamily\footnotesize,
    breaklines=true,
    breakatwhitespace=false,
    columns=fullflexible,
    keepspaces=true,
    frame=none,
    postbreak=\mbox{\textcolor{gray}{$\hookrightarrow$}\space},
]
{
  "popularity": "popular | non-popular"
}
\end{lstlisting}

\end{promptbox}

\subsection{Examples of High-level Guidance Agent}
\label{appendix:prompt_of_guidance_agent}

\begin{promptbox}[title=Prompt for Question Graph Parsing]
\small

\textbf{Role: Question Graph Parsing Expert} \\
Parse the given \texttt{(question, answer, reasoning\_chain)} into a strict Question Graph.

\textbf{Graph Definition}
\begin{itemize}\itemsep0pt
    \item \textit{Node}: a set of entities or attribute values.
    \item \textit{Edge}: one of exactly three operations.
    \begin{itemize}\itemsep0pt
        \item \texttt{projection} ($\pi_r$): traverse one step along relation $r$, either forward or backward.
        \item \texttt{intersection} ($\cap$): take the intersection of multiple entity sets.
        \item \texttt{complement} ($\neg$): take the complement of an entity set.
    \end{itemize}
\end{itemize}

\textbf{Core Principles}
\begin{enumerate}\itemsep0pt
    \item \textit{Faithful to the literal semantics of the question}: the \texttt{reasoning\_chain} serves only as a reference for entity and relation names; do not use it to forcibly connect subgraphs that have no logical dependency in the question itself.
    \item \textit{Local connectivity}: even when an isolated subgraph functions as mere ``background filler,'' its internal edges must still be built according to the literal logic expressed in the question.
\end{enumerate}

\textbf{Node Fields}
\begin{itemize}\itemsep0pt
    \item \texttt{node\_id}: starts from \texttt{n0}.
    \item \texttt{label}: semantic description of the node.
    \item \texttt{known\_entities}: entities explicitly given in the question; otherwise \texttt{null}.
    \item \texttt{reference\_entities}: real entity names not mentioned in the question but provided by the reasoning\_chain; otherwise \texttt{null}.
    \item \texttt{is\_root}: whether this node is the answer node.
\end{itemize}

\textbf{Input}
\begin{itemize}\itemsep0pt
    \item \texttt{question}: \{question\}
    \item \texttt{answer}: \{answer\}
    \item \texttt{reasoning\_chain}: \{reasoning\_chain\}
\end{itemize}

\textbf{Output Format} \\
Output only the following JSON, with no additional text:

\begin{lstlisting}[
    basicstyle=\ttfamily\footnotesize,
    breaklines=true,
    breakatwhitespace=false,
    columns=fullflexible,
    keepspaces=true,
    frame=none,
    postbreak=\mbox{\textcolor{gray}{$\hookrightarrow$}\space},
]
{
  "question": "...",
  "answer": "...",
  "root_node_id": "n_x",
  "nodes": [
    {
      "node_id": "n0", 
      "label": "...", 
      "known_entities": [...] or null,
      "reference_entities": [...] or null, 
      "is_root": false}
  ],
  "edges": [
    {
      "edge_id": "e0", 
      "source_ids": ["n0"], 
      "target_id": "n1", 
      "op": "projection | intersection | complement", 
      "relation": "...", 
      "direction": "forward | backward"}
  ]
}
\end{lstlisting}

\end{promptbox}

\begin{promptbox}[title=Prompt for QA Synthesis Direction]
\small

\textbf{Role: QA Synthesis Direction} \\
Given the parsed Question Graph, propose \textbf{one} concrete expansion suggestion that makes the question more complex and adversarial.

\textbf{Input}
\begin{itemize}\itemsep0pt
    \item \texttt{question}: \{question\}
    \item \texttt{answer}: \{answer\}
    \item \texttt{graph\_json}: \{graph\_json\}
\end{itemize}

\textbf{Allowed Operations} (only the following three edge types may be added)
\begin{itemize}\itemsep0pt
    \item \texttt{projection}: traverse one step along a relation.
    \item \texttt{intersection}: add an additional condition that must be jointly satisfied.
    \item \texttt{complement}: add an exclusionary constraint.
\end{itemize}

\textbf{Candidate Strategies}
\begin{itemize}\itemsep0pt
    \item \textit{Strategy A (Complement)}: attach a ``never / excluding \ldots'' constraint to a node, producing a hard negative that retrieval systems struggle to handle.
    \item \textit{Strategy B (Inverse Projection)}: flip a forward projection into its inverse (effect $\rightarrow$ cause), which is inherently harder to reason about than forward inference.
    \item \textit{Strategy C (Intersection Expansion)}: add an intersection branch on an intermediate node. Placing it on an early node (left-deep tree) or on a node close to the answer (right-deep tree) modulates how the reasoning difficulty is distributed along the path.
\end{itemize}

\textbf{Output Format} \\
Output only the following JSON, with no additional text:

\begin{lstlisting}[
    basicstyle=\ttfamily\footnotesize,
    breaklines=true,
    breakatwhitespace=false,
    columns=fullflexible,
    keepspaces=true,
    frame=none,
    postbreak=\mbox{\textcolor{gray}{$\hookrightarrow$}\space},
]
{
  "expansion_advice": {
    "target_text_span": "the exact text span of the target entity copied verbatim from the original question",
    
    "suggested_operation": "complement | projection | intersection",
    
    "selected_strategy": "Complement | Inverse Projection | Intersection Expansion",
    
    "reasoning": "why this operation and strategy are chosen, and how they increase the graph's complexity and reasoning difficulty",
    
    "semantic_suggestion": "a natural-language description of the concrete modification to apply"
  }
}
\end{lstlisting}

\end{promptbox}

\section{Low Quality QA pairs filtering}
\label{appendix:qa_filtering}

\begin{promptbox}[title=Prompt for QA Data Quality Filtering]
\small

\textbf{Role: QA Data Quality Filtering} \\
Inspect the given QA pair and output only ``pass'' or ``fail''.

\textbf{Input}
\begin{itemize}\itemsep0pt
    \item \texttt{question}: \{question\}
    \item \texttt{answer}: \{answer\}
\end{itemize}

\textbf{Inspection Criteria} (failing any single item results in ``fail'')

\textbf{1. Completeness}
\begin{itemize}\itemsep0pt
    \item Both question and answer are non-empty, non-whitespace, and free of truncation or garbled text.
\end{itemize}

\textbf{2. Question Language Quality}
\begin{itemize}\itemsep0pt
    \item Fluent and natural, consistent with how a human would naturally ask, with no grammatical errors or signs of machine-generated patchwork.
    \item Sentence structure is clear, avoiding excessive nested modifiers.
    \item Semantically unique and unambiguous, with no unclear references, vague scope, or multiple valid interpretations.
    \item Forms a complete interrogative sentence with an explicit target of inquiry.
\end{itemize}

\textbf{3. Answer Quality}
\begin{itemize}\itemsep0pt
    \item Concise and definite, easy to evaluate automatically (e.g., a specific entity, number, or date).
    \item Semantically unique, with no equally valid alternative answers.
    \item Free of ambiguous expressions or vague qualifiers.
\end{itemize}

\textbf{Output Format} \\
Output only the following JSON, with no additional text or code-block markers:

\begin{lstlisting}[
    basicstyle=\ttfamily\footnotesize,
    breaklines=true,
    breakatwhitespace=false,
    columns=fullflexible,
    keepspaces=true,
    frame=none,
    postbreak=\mbox{\textcolor{gray}{$\hookrightarrow$}\space},
]
{
  "thinking": "brief reasoning",
  
  "result": "pass | fail"
}
\end{lstlisting}

\end{promptbox}

\section{Human Validation}
\subsection{Human Analyses on Data Quality}
\label{appendix:data_quality}

Two data experts from the author team, proficient in both English and Chinese, are involved in the human analyses.
For (1) whether each $\epsilon_i \in \mathcal{E}$ is correct: each expert verifies whether $\epsilon_i$ is consistent with or can be inferred from its source web pages.
For (2) whether each synthesized question is consistent with its corresponding $\mathcal{E}$ and is unambiguous; and (3) whether each answer can be inferred by the corresponding $\mathcal{E}$, the data experts also give their judgments independently.
In cases where the two experts have differing judgments on a specific issue, the final decision will be reached through discussion.

\subsection{Human Baseline Details}
\label{appendix:human_baseline}

Six evaluators proficient in both English and Chinese all have NLP/LLM backgrounds, hold a Master’s or PhD degree, and had no prior exposure to the QA samples in EvoBrowseComp.
Two data experts judge the correctness of the answers: a QA is marked as ``solved within 2h'' if the answer is judged correct; otherwise the evaluator keeps working, and if the QA remains unsolved after 2h, it is marked as ``gave up after 2h''.

\section{Prompt of LLM-as-a-judge}
\label{appendix:judge_prompt}

\begin{promptbox}[title=Prompt for LLM-as-a-Judge]
\small

You are a general artificial intelligence assistant. Based on the \texttt{[Correct Answer]} provided below, judge whether the \texttt{[Response]} to the \texttt{[Original Question]} below is correct.

\vspace{0.5em}
\texttt{[Original Question]}: \{question\}

\vspace{0.3em}
\texttt{[Correct Answer]}: \{correct\_answer\}

\vspace{0.3em}
\texttt{[Response]}: \{response\}

\vspace{0.5em}
Your judgment must follow the following format and standards:

\vspace{0.5em}
\textbf{Final Answer}: The final accurate answer extracted from the \texttt{[Response]}. If there is no clear final answer in the \texttt{[Response]}, fill in \texttt{'None'}.

\vspace{0.3em}
\textbf{Explanation}: Explain why the \texttt{[Final Answer]} is correct or incorrect based on the \texttt{[Correct Answer]}. Only focus on whether there is a substantial difference between the \texttt{[Final Answer]} and the \texttt{[Correct Answer]}; do not comment on the background of the question, do not attempt to re-solve the problem, do not defend any answer different from the \texttt{[Correct Answer]}, and only focus on judging whether the answers are consistent.

\vspace{0.3em}
\textbf{Conclusion}: If the \texttt{[Final Answer]} is consistent with the \texttt{[Correct Answer]} provided above, or within an acceptable small error range for numerical questions, fill in \texttt{'Correct'}; otherwise (i.e., in case of any inconsistency, ambiguity, inequivalence, or incorrect extracted answer), fill in \texttt{'Incorrect'}.

\end{promptbox}

\begin{table}[t]
\centering
\resizebox{0.35\textwidth}{!}
{
\begin{tabular}{lc}
\toprule[1pt]
\multicolumn{1}{c}{Model} & \multicolumn{1}{c}{Spearman} \\ \midrule[1pt]
GPT-4.1      &   0.830    \\
DeepSeek-V4-Flash-Chat      &   0.812    \\
DeepSeek-V4-Flash-Max      &  0.828     \\
DeepSeek-V3.2-Chat      &  0.824     \\
DeepSeek-V3.2-Think      &  0.803     \\
Kimi-K2.6-Chat      &   0.730    \\
Kimi-K2.6-Think      &   0.843    \\
GLM-5-Chat      &   \textbf{0.864}    \\
GLM-5-Think     &   0.847    \\
\bottomrule[1pt]
\end{tabular}
}
\caption{The Spearman correlation between LLM judge and human judge.}
\label{table:llm_judge}
\end{table}

\section{Selection of the Judge Model}
\label{appendix:judge_model}

To select a reliable judge model in model evaluation, we randomly select 800 model predictions from the main experiments (Section~\ref{subsec:3.1}), and respectively use the following cutting-edge LLMs as the judge models: GPT-4.1~\cite{o3}, DeepSeek-V4-Flash-Chat, DeepSeek-V4-Flash-Max~\cite{deepseekai2026deepseekv4}, DeepSeek-V3.2-Chat, DeepSeek-V3.2-Think~\cite{deepseekv3.2}, Kimi-K2.6-Chat, Kimi-K2.6-Think~\cite{kimik26}, GLM-5-Chat and GLM-5-Think~\cite{zeng2026glm}.
To figure out their correlation with humans, we also manually annotate whether each model prediction is correct or not.
Similar to the human analyses (Appendix~\ref{appendix:data_quality}), two data experts participate in the human annotation process, independently making their assessments and subsequently discussing any discrepancies in their judgments.
As shown in Table~\ref{table:llm_judge}, GLM-5-Chat achieves the best correlation with human judgments in terms of Spearman correlation~\cite{zar2005spearman}.

\section{Average Number of Tool Calls}
\label{appendix:tool_calls}

To further investigate whether EvoBrowseComp meaningfully evaluates multi-hop agentic search, we report the average number of tool calls per question for each model in Table~\ref{tab:tool_calls}, where ``search'' and ``visit'' denote the call counts for the search and visit tools, respectively.
Across all models, solving a single question requires a substantial number of tool calls (\emph{e.g.}, over 20 search calls for DeepSeek-V4 series), reflecting the multi-hop nature of EvoBrowseComp and confirming that it effectively evaluates agentic search capabilities.

\begin{table}[t]
\centering
\resizebox{0.48\textwidth}{!}
{
\begin{tabular}{lcccc}
\toprule[1pt]
\multicolumn{1}{c}{\multirow{2}{*}{Model}} & \multicolumn{2}{c}{EN} & \multicolumn{2}{c}{ZH} \\
\cmidrule(lr){2-3} \cmidrule(lr){4-5}
 & Search & Visit & Search & Visit \\ \midrule[1pt]
GLM-5                     & 15.5 & 11.2 & 17.3 & 9.3 \\
Kimi-K2.6                 & 16.6 & 2.9  & 19.5 & 2.0 \\
Qwen3.5-397B-A17B-FP8     & 10.7 & 4.8  & 10.1 & 4.3 \\
Qwen3.5-122B-A10B         & 13.8 & 5.0  & 11.8 & 4.0 \\
Qwen3.5-35B               & 21.4 & 7.8  & 24.6 & 8.2 \\
Qwen3.5-27B               & 12.7 & 2.0  & 11.8 & 1.7 \\
Qwen3-235B-A22B-Thinking  & 11.6 & 2.3  & 12.4 & 1.8 \\
DeepSeek-V3.2             & 23.8 & 8.9  & 24.7 & 4.9 \\
DeepSeek-V4-Flash         & 24.2 & 12.8 & 30.6 & 6.9 \\
DeepSeek-V4-Pro           & 24.8 & 11.0 & 29.2 & 6.4 \\
Claude-Opus-4.6           & 14.4 & 8.8  & 17.8 & 8.6 \\ \bottomrule[1pt]
\end{tabular}
}
\caption{Average number of tool calls per question for each model on EvoBrowseComp.}
\label{tab:tool_calls}
\end{table}
\end{document}